\pdfoutput=1
\documentclass[11pt]{article}

\usepackage[preprint]{acl}

\usepackage{times}
\usepackage{latexsym}

\usepackage[T1]{fontenc}
\usepackage[utf8]{inputenc}

\usepackage{microtype}

\usepackage{inconsolata}

\usepackage{graphicx}
\usepackage{float}
\usepackage{hyperref}       % hyperlinks
\usepackage{url}            % simple URL typesetting
\usepackage{booktabs}       % professional-quality tables
\usepackage{amsfonts}       % blackboard math symbols
\usepackage{nicefrac}       % compact symbols for 1/2, etc.
\usepackage{xcolor}         % colors
\usepackage{caption}
\usepackage{subfigure}
\usepackage{amsmath,amsfonts}
\usepackage{multirow}
\usepackage{algorithmic}
\usepackage{enumitem}
\usepackage{graphicx}
\usepackage{color}
\usepackage[linesnumbered, ruled, vlined]{algorithm2e}
\usepackage{changes}%add [final] before {changes}, print the 
\usepackage{wrapfig}

\newcommand{\toolname}{SSDAU}

\title{SSDAU: Structured Semantic Data Augmentation for Joint Entity and Relation Extraction}

\author{
Jiawei He$^{1,2,\dagger}$, Mengyu Shi$^{1,\dagger}$, Jiawei Liu$^{1}$, Dong Sun$^{2}$,\\
Chunrong Fang$^{1,*}$, Xikai Yang$^{2,*}$, Zhijie Wang$^{3,*}$, Lei Ma$^{3,4}$, Zhenyu Chen$^{1,*}$\\[0.5em]
\normalsize
$^{1}$State Key Laboratory for Novel Software Technology, Nanjing University, Nanjing, China\\
$^{2}$Amap, Alibaba Group, China\\
$^{3}$University of Alberta, Edmonton, Canada\\
$^{4}$The University of Tokyo, Tokyo, Japan\\[0.3em]
$^{\dagger}$Contribute equally \quad
$^{*}$Corresponding author
}

\begin{document}

\maketitle

\begin{abstract}
Joint Entity and Relation Extraction (JERE) is highly sensitive to training data quality, making data augmentation a natural way to improve generalization. 
However, existing augmentation methods often weaken entity relevance and disrupt semantic structure, limiting their effectiveness for JERE. 
In this paper, we propose \textbf{Structured Semantic Data Augmentation (SSDAU)}, a method designed to preserve triple-aware semantic structure during augmentation.
SSDAU segments text by entity labels, captures semantic features through context-aware encoding, and restructures entity semantics to generate augmented data.
To distinguish semantically similar entities, SSDAU combines contextualized embeddings with traditional similarity scores.
To reduce topic inconsistency, we apply BERTopic-based filtering to remove irrelevant augmentations.
We evaluate SSDAU on datasets with different annotation types and compare its performance on five representative JERE models against seven popular augmentation baselines.
Experiments show that SSDAU generates semantically consistent data, is more robust to ambiguity than non-LLM methods (8.95\% vs. 23.58\% average relative F1 decrease), and significantly outperforms strong alternatives in most settings.
  
  % The abstract paragraph should be indented \nicefrac{1}{2}~inch (3~picas) on
  % both the left- and right-hand margins. Use 10~point type, with a vertical
  % spacing (leading) of 11~points.  The word \textbf{Abstract} must be centered,
  % bold, and in point size 12. Two line spaces precede the abstract. The abstract
  % must be limited to one paragraph.
\end{abstract}

% \vspace{-0.4cm}
\section{Introduction}
Joint Entity and Relation Extraction (JERE) is a fundamental information extraction task with applications in information retrieval~\citep{lin2020joint}, question answering~\citep{abdelaziz2021semantic}, and text summarization~\citep{zhong2020extractive}. Its performance depends heavily on the quality and diversity of the training data. Data augmentation is therefore a natural way to improve robustness and generalization, especially when training data are noisy, partially annotated, or distributionally limited. For example, methods such as MixUp~\citep{cheng2020advaug} and back-translation~\citep{xie2020unsupervised} generate new training instances through controlled perturbations of the original text.

However, applying generic augmentation techniques to JERE is non-trivial. Because JERE predicts structured triples rather than sentence-level labels, perturbing the original text may weaken entity relevance, alter relation cues, or break the dependency structure among entities and relations~\citep{kambhatla2022cipherdaug}. This issue is particularly important in the presence of overlapping relations and tightly coupled entity--relation decisions~\citep{liu2020document}. As a result, augmented samples that are fluent at the surface level may still be harmful for JERE training if they violate triple-level semantic constraints.

To address this issue, we propose Structured Semantic Data Augmentation ({\toolname}), a three-stage pipeline designed to preserve triple-aware semantic structure during augmentation. Instead of perturbing complete sentences, {\toolname} segments each sentence by entity labels, matches segments under shared semantic constraints, and reconstructs new training instances through structure-consistent replacement. To better distinguish semantically similar but non-equivalent entities, {\toolname} combines contextualized embeddings with traditional similarity scores. Because structured replacement alone can still introduce noisy or topic-inconsistent samples, {\toolname} further applies topic-aware consistency filtering to remove candidate augmentations incompatible with the original triple semantics.

\begin{figure*}[t]
  \centering
  \includegraphics[width=0.8\linewidth]{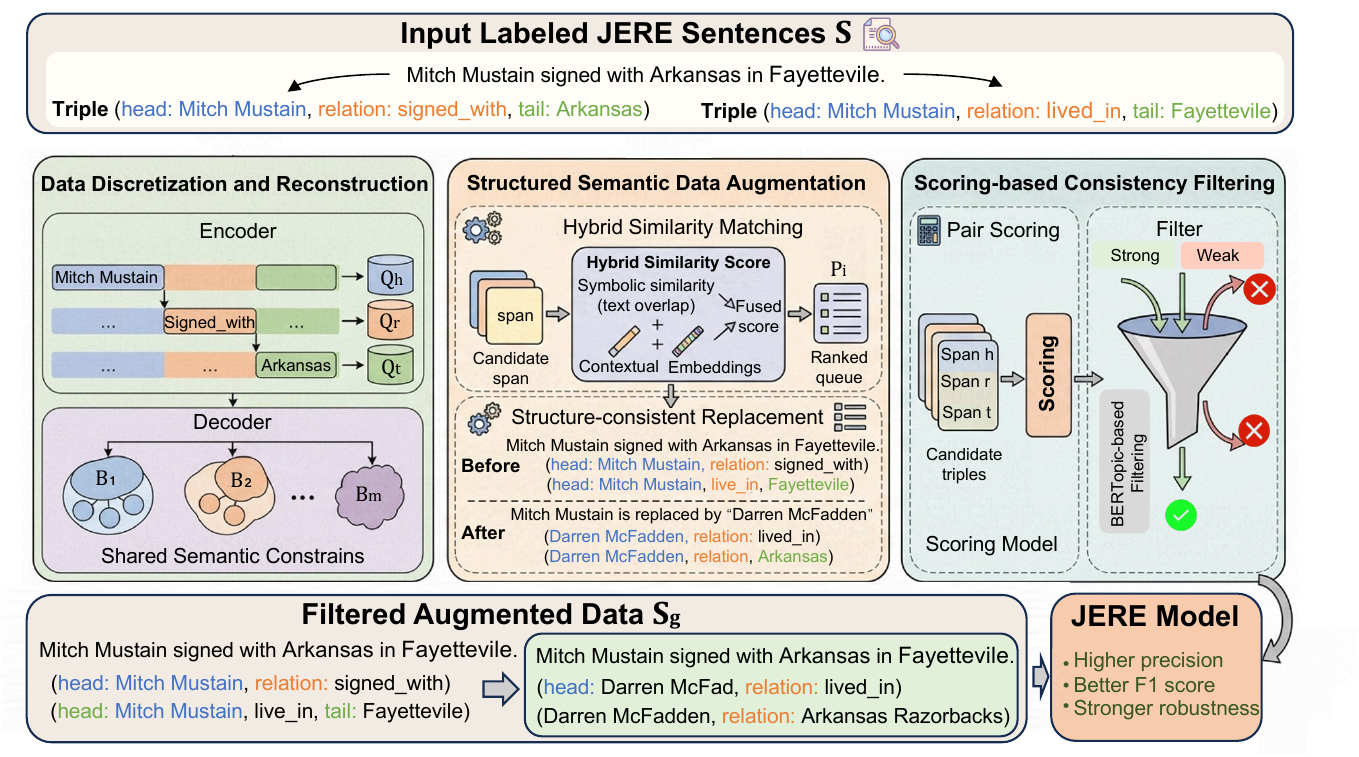}
%\vspace{-0.3cm}
%\captionsetup{font={small}}
\caption{\textbf{Overview of {\toolname}}. Given labeled JERE sentences $S$, {\toolname} first applies \textit{\textbf{Data Discretization and Reconstruction}} to extract head, relation, and tail spans and group them under shared semantic constraints. It then performs \textit{\textbf{Structured Semantic Data Augmentation}} by ranking candidate spans with hybrid similarity scores and generating new instances through structure-consistent replacement. Finally, \textit{\textbf{Scoring-based Consistency Filtering}} evaluates candidate triples with pair scoring and BERTopic-based filtering, retaining only coherent augmentations. The filtered data $S_{g}$ are used to train a JERE model with higher precision, better F1, and stronger robustness.
}

  \label{flows}
\end{figure*}

To evaluate {\toolname}, we compare it with seven augmentation baselines on four benchmark settings and five representative JERE models. Experimental results show that {\toolname} improves average performance in most settings and remains stable under semantic perturbation. The ablation results further show that both structure-aware replacement and consistency filtering contribute to the final performance, with filtering playing an important role in controlling augmentation noise. Under the clean setting, {\toolname} achieves the best average precision and F1 among the compared methods.

% \vspace{-0.2cm}
\section{Related Work}
% \vspace{-0.2cm}
% \vspace{-0.4cm}
\paragraph{Data Augmentation} 
Traditional augmentation methods, such as synonym replacement~\citep{wei2019eda} and back-translation~\citep{xie2020unsupervised}, often improve lexical diversity but may disrupt syntactic dependencies or triple consistency crucial for JERE~\citep{kambhatla2022cipherdaug}. Recent LLM-based methods improve fluency and controllability~\citep{nawara2025comprehensive}, but can still introduce structurally invalid or semantically mismatched triples in IE-oriented augmentation~\citep{fang2025zero}. Recent graph-guided, retrieval-based, or syntax-aware methods also aim to better preserve structure~\citep{tao2025retrieval}, although they may require additional resources or offer limited control at the entity level. In contrast, {\toolname} uses a lightweight structured replacement pipeline with explicit semantic constraints and a post-hoc consistency filter to preserve triple topology without relying on autoregressive generation.

\paragraph{Semantic Consistency.}
Ensuring semantic fidelity in augmented data is crucial~\citep{harkous2020have}. However, cosine similarity on contextual embeddings~\citep{kenton2019bert} often captures lexical or surface-level closeness while missing structural incompatibility between entities. Recent work mitigates this with \textit{hybrid matching}~\citep{yang2026sparse} and \textit{topic-aware filtering}~\citep{hofstatter2021efficiently}, but these are typically applied only after candidate generation. In contrast, \toolname incorporates hybrid similarity scoring and BERTopic-based consistency filtering directly into the augmentation loop, steering generation toward semantically relevant, topically coherent, and structurally valid instances.

% \vspace{-0.15cm}
\section{Method}

% \vspace{-0.5cm}

In this section, we first formalize the problem setting and then present the three core components of {\toolname}: (1) data discretization and reconstruction, (2) structured semantic data augmentation, and (3) scoring-based consistency filtering. Figure~\ref{flows} overviews the framework. For clarity, we use $n$ to denote the maximum sentence length and $K$ to denote the number of predefined relation types.
% The complete process is also illustrated in Algorithm~\ref{alg:algorithm1}.

% provide an overview of our proposed data augmentation method. Figure \ref{flows} depicts the framework of SSDAU. We first define the tasks and then describe the three essential components of SSDUA, including i) text segmentation and reconstruction, ii) data augmentation with structured semantic text matching, and iii) consistent filtering of scoring classifiers. The complete process is illustrated in Algorithm \ref{alg:algorithm1}.

% \vspace{-0.1cm}
\subsection{Preliminaries}
% \vspace{-0.1cm}
Given a set of sentences $S=\{s_{1},s_{2},...,s_{N}\}$, where each sentence contains up to $n$ tokens, and a predefined relation set $R=\{r_{1},r_{2},...,r_{K}\}$, we extract triples $T=\{(h_{i},r_{i},t_{i})\}_{i=1}^{M}$ from $S$, where $h_i$ and $t_i$ denote the head and tail entities, respectively. Following common JERE formulations, we represent triple labels with a tensor in $\mathbb{R}^{n \times K \times n}$.

Since triplets are the core output format of JERE, we use the triplet as the basic unit of data augmentation and partition text according to the triplet to obtain three series of text collections. To preserve the contextual semantics of the segmented text, we keep the contextual token $l$ of each segment and record each cut point location $p$. For clarity, we use $\varepsilon$ to denote the minimum hybrid similarity required for candidate replacement. In the case study, $\varepsilon_{1}$ and $\varepsilon_{2}$ denote the thresholds for entity and relation replacement, respectively. We use $\nu \in [0,1]$ to denote a normalized syntactic coherence score, where larger values indicate better agreement with the source syntactic pattern.

% \vspace{-0.2cm}
\subsection{Data Discretization and Reconstruction}
% \vspace{-0.1cm}

\paragraph{Encoder}
We use triples as the basic augmentation unit to avoid noise from direct perturbation. Given a sentence $s_i$, the encoder $E$ identifies the head, relation, and tail spans ($q_{h_i}, q_{r_i}, q_{t_i}$) based on the corresponding triple tags ($\rho_{h_i}, \rho_{r_i}, \rho_{t_i}$), records their context tokens and boundary positions, and outputs three text collections: $Q_h$, $Q_t$, and $Q_r$.

\begin{figure}[t]
  \centering
  \includegraphics[width=0.95\linewidth]{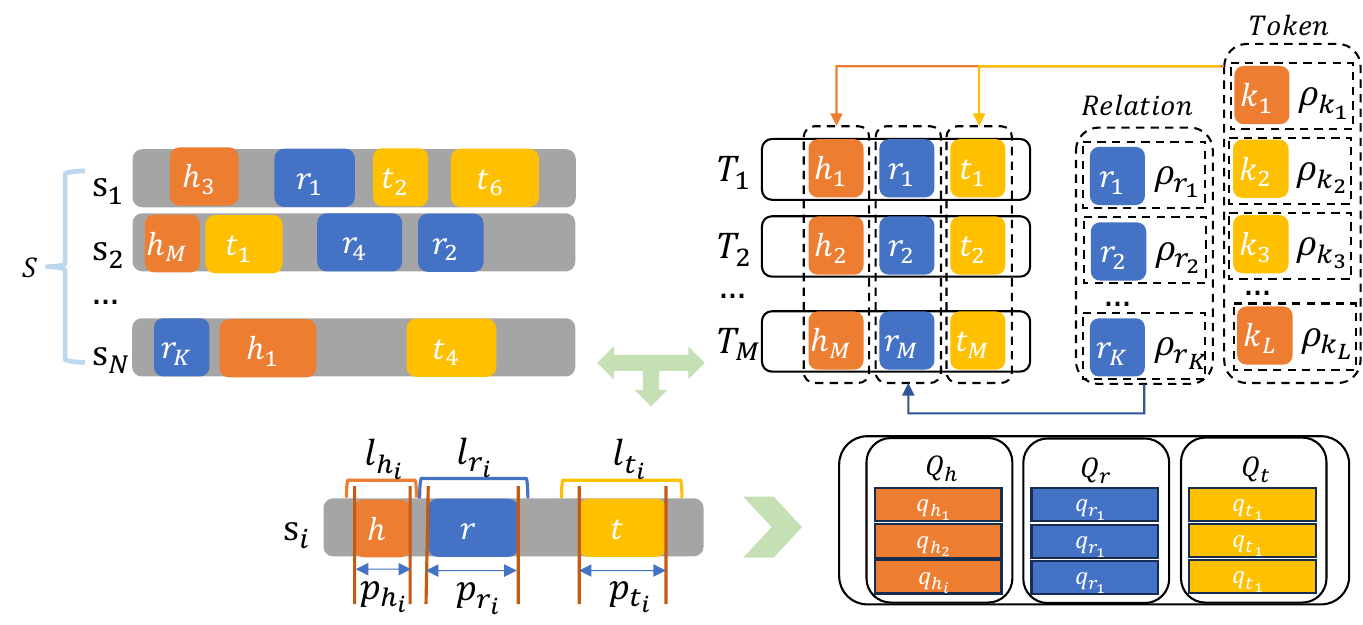}
%\vspace{-0.3cm}
%\captionsetup{font={small}}
\caption{The structure of our feature-based encoder.}

  \label{encoder}
\end{figure}

\paragraph{Decoder} We then design a similarity-based text matching decoder $D$, which takes $(Q_h, Q_t, Q_r)$ as input. The decoder partitions the text collections according to relation types and label types into a set of groups $B=\{B_1, B_2, \dots, B_m\}$, where each group contains text segments with the same semantic constraints.

% similarity form based on $K$ relation types and $M$ triplet labels in the sentence set $S$., and use it as the basis for designing a form similarity-based text matching decoder $D$. The input of decoder $D$ is $(Q_{h}, Q_{t}, Q_{s})$, and it divides the text collections according to the relation types and label types  to get $LKL$ groups text library $B=\{B_{1},B_{2},...,B_{LKL}\}$ with the same relation type and the same label.

% \vspace{-0.1cm}
\subsection{Structured Semantic Data Augmentation}
\label{sec:augmentation}
% \vspace{-0.1cm}
\paragraph{Discrete Text Matching}
We design a text matcher based on the semantic similarity tool \emph{Similarities}~\citep{Xu_Similarities_Compute_similarity} to align the decoder output. 
A text block $b$ in an output group $B_{i} = \{b_{1}, b_{2}, \dots, b_{j}\}$ stores the text span $q$, context tokens $l$, label type $\rho$, and segmentation position $p$. We perform matching across text blocks from compatible groups $B_i$ by combining semantic, syntactic, lexical, and context-token similarity signals. To better distinguish semantically similar but non-equivalent entities, we additionally encode each candidate with a pretrained BERT encoder and compute a contextual similarity score from the pooled representation. The final hybrid similarity is the normalized weighted combination of the original similarity score and the contextual similarity score, and lies in $[0,1]$. We then insert matched candidates into a priority queue sorted in descending order of hybrid similarity. For each group $B_i$, this yields a similarity-based priority queue $P_i$.

% \vspace{-0.2cm}
\paragraph{Data Augmentation}
After similarity matching, we retain only candidates in the priority queue $P_i$ whose hybrid similarity is no smaller than the threshold $\varepsilon$. For each retained candidate, we replace the corresponding head, tail, or relation span in the source sentence according to the recorded segmentation position $p$, while preserving the remaining sentence context and the triple label configuration. The resulting augmented instances are appended to the original training set rather than replacing the original samples.

\subsection{Scoring-based Consistency Filtering}
To further improve the quality of the augmented data, we employ a BERTopic model to identify topic-relevant terms and filter out augmented data associated with irrelevant topics, thereby improving topic coherence.
% To this end, we design a score-based BertTopic text filter.

First, we extract all entities and relations from the text. Then, we encode the tokens using BERT~\citep{kenton2019bert}, obtaining the corresponding entity tokens ${l_{1}, l_{2}, \dots, l_{L}}$. Next, we combine entities and relations in the form of $(l_{h}, r, l_{t})$ and perform triplet extraction using joint entity and relation extraction~\citep{OneRel}. Finally, we compute a contextualized representation for each head--tail entity pair.
The pair representation is defined as:
\begin{equation}
\begin{split}
    \mathbf{e}_{ht}=\phi\!\left(W[l_{h};l_{t}]^{T}+b\right)
\end{split}
\end{equation}
where $l_h$ and $l_t$ denote the contextual representations of the head and tail entities, respectively, and $\mathbf{e}_{ht}$ is the resulting head--tail pair representation. 
Here, $W \in \mathbb{R}^{d_e \times 2d}$ and $b \in \mathbb{R}^{d_e}$ are trainable parameters, $[;]$ denotes concatenation, and $\phi(\cdot)$ is the ReLU activation function. In our implementation, the pair-scoring module is initialized using statistics derived from pretrained contextual representations produced by the BERT encoder, and all parameters are then optimized during filtering. Our ablation studies confirm that this pretrained-context initialization outperforms random and zero initialization in our setting.

We then project the pair representation to relation scores with a trainable relation matrix $R \in \mathbb{R}^{d_e \times K}$. The score vector is defined as follows:
\begin{equation}
% \scriptsize
% \begin{split}
    \mathbf{v}_{ht}=R^{T}\,\mathrm{drop}(\mathbf{e}_{ht})
% \end{split}
\end{equation}
where $\mathbf{v}_{ht}\in\mathbb{R}^{K}$ is the relation score vector for the head--tail pair, and $\mathrm{drop}(\cdot)$ denotes dropout.

We then apply a softmax over $\mathbf{v}_{ht}$ to predict the corresponding relation label. To assess whether an augmented instance remains compatible with the source triple semantics, we compute an average triple-level consistency loss:
\begin{equation}
\small
  \begin{aligned}
    \zeta_{\text{triple}} = - \frac{1}{n \times K \times n}\sum_{i,j,k}\log P\!\left(g_{(l_i,r_k,l_j)} \mid S\right),
  \end{aligned}
\end{equation}
where $g_{(l_i,r_k,l_j)}$ is the gold relation tag from the original annotation, and $P(g_{(l_i,r_k,l_j)} \mid S)$ is the predicted probability assigned to that tag for sentence $S$. Lower $\zeta_{\text{triple}}$ indicates better semantic consistency with the source annotation. For filtering, we therefore convert this loss into a consistency score by ranking candidates in ascending order of $\zeta_{\text{triple}}$ (equivalently, descending order of $-\zeta_{\text{triple}}$), and discard augmented instances whose consistency is below the validation-selected threshold. In this way, topic-aware consistency filtering reduces error propagation by removing candidate replacements incompatible with the source semantics.

% \vspace{-0.2cm}
\section{Experiment}
% \vspace{-0.15cm}

\subsection{Experimental Setup}
\paragraph{Baseline}
We compare {\toolname} with seven widely used data augmentation methods: \textit{word substitution} (WS) \citep{wei2019eda}, \textit{back translation} (BT) \citep{xie2020unsupervised}, \textit{noise introduction} (NI) \citep{xie2017data}, \textit{same-tag semantic noise} (SSN) \citep{yan2019data}, \textit{generative models} (GM) \citep{hou2021c2c}, Mixup \citep{hu2019learning}, and ChatIE \citep{wei2023chatie}. We implement all baselines under the same training and evaluation pipeline and select their controllable thresholds or augmentation settings on the validation split whenever applicable. ChatIE and the additional ``LLM Aug.'' result are included as reference generation-based baselines under our unified setting rather than as a claim of current frontier LLM performance.

\begin{table}[t]
\caption{Augmented sample counts under different similarity thresholds. We select $\varepsilon=0.7$ on the validation set for all main experiments.} 
\label{Epsilon}
\centering
% \tabcolsep=0.1cm
% \vspace{0.4em} 
\centering%  xiaowu
% \resizebox{65mm}{34mm}{
\footnotesize
\setlength{\tabcolsep}{3pt}
\begin{tabular}{lcrrrr}%{\textwidth}
%\toprule[0.001pt]
\toprule
% \rule{0pt}{13pt} %此命令用于更改行高，
% \vspace{0.1cm}
 Dataset &  $\varepsilon$ &  \multicolumn{1}{c}{Head} &  \multicolumn{1}{c}{Relation} &  \multicolumn{1}{c}{Tail} &  \multicolumn{1}{c}{Sum.}\\
\midrule
\multirow{5}{*}{$NYT^{\ast}$} & 0.5 $\sim$ 0.6 & 15,062 & 243 & 11,300 & 26,605\\
   & 0.6 $\sim$ 0.7 & 9,439 & 38 & 4,631 & 14,108\\
   & 0.7 $\sim$ 0.8 & 1,825 & 19 & 1,365 & 3,209\\
   & 0.8 $\sim$ 0.9 & 2,927 & 0 & 1,137 & 4,064\\
   & 0.9 $\sim$ 1.0 & 960 & 0 & 1,546 & 2,506\\
% \hline
 % & Sum. & 30213 & 300 & 19979 & 50492\\
\midrule
\multirow{5}{*}{$WebNLG^{\ast}$} & 0.5 $\sim$ 0.6 & 7,082 & 2,742 & 8,116 & 17,940\\
   & 0.6 $\sim$ 0.7 & 3,933 & 1,946 & 5,342 & 11,221\\
   & 0.7 $\sim$ 0.8 & 2,049 & 2,162 & 1,557 & 5,768\\
   & 0.8 $\sim$ 0.9 & 814 & 2,005 & 1,021 & 3,840\\
   & 0.9 $\sim$ 1.0 & 5,463 & 890 & 2,929 & 9,282\\
% \hline
 % & Sum. & 19341 & 9745 & 18965 & 48051\\
\midrule
   \multirow{5}{*}{$NYT$} & 0.5 $\sim$ 0.6 & 13,507 & 234 & 10,076 & 23,817\\
   & 0.6 $\sim$ 0.7 & 7,721 & 36 & 4,063 & 11,820\\
   & 0.7 $\sim$ 0.8 & 4,922 & 13 & 1,588 & 6,523\\
   & 0.8 $\sim$ 0.9 & 2,198 & 0 & 1,140 & 3,338\\
   & 0.9 $\sim$ 1.0 & 3,700 & 0 & 1,051 & 4,751\\
% \hline
 % & Sum. & 32048 & 283 & 17918 & 50249\\
\midrule
\multirow{5}{*}{$WebNLG$} & 0.5 $\sim$ 0.6 & 4,023 & 3,186 & 6,028 & 13,237\\
   & 0.6 $\sim$ 0.7 & 2,673 & 2,009 & 4,445 & 9,127\\
   & 0.7 $\sim$ 0.8 & 968 & 1,345 & 1,123 & 3,436\\
   & 0.8 $\sim$ 0.9 & 309 & 919 & 923 & 2,151\\
   & 0.9 $\sim$ 1.0 & 3,019 & 444 & 6,935 & 10,398\\
% \hline
 % & Sum. & 10992 & 7903 & 19454 & 38349\\
\bottomrule
\end{tabular}

% }
\end{table}

\begin{table*}[t]
\caption{Comparison with baselines under clean and semantically perturbed settings. All values are averaged over five random seeds; no variance estimates are shown due to space constraints.}
\label{combined_results}
\footnotesize
\centering
\setlength{\tabcolsep}{1.8pt}
\begin{tabular}{llcccccccccccc}
\toprule
    \multirow{2}{*}{Method} & \multirow{2}{*}{Quality} & \multicolumn{3}{c}{NYT$^{\ast}$} & \multicolumn{3}{c}{WebNLG$^{\ast}$} & \multicolumn{3}{c}{NYT} & \multicolumn{3}{c}{WebNLG} \\
    \cmidrule(r){3-5}
    \cmidrule(r){6-8}
    \cmidrule(r){9-11}
    \cmidrule(r){12-14}
    & & Prec. & F1 & IoU & Prec. & F1 & IoU & Prec. & F1 & IoU & Prec. & F1 & IoU \\
\midrule
\multirow{1}{*}{Original} & -- & 90.17 & 91.45 & 84.24 & 90.62 & 90.25 & 82.23 & 92.83 & 92.17 & 85.47 & 90.66 & 89.08 & 80.31 \\
\midrule
\multirow{2}{*}{WS} & Clean & 88.82 & 88.98 & 80.16 & 91.47 & 91.51 & 84.35 & 89.91 & 89.61 & 81.17 & 89.66 & 88.88 & 79.98 \\
 & Perturbed & 75.60 & 76.30 & 61.80 & 77.10 & 75.20 & 60.30 & 78.20 & 71.40 & 55.60 & 75.80 & 69.50 & 53.20 \\
\midrule
\multirow{2}{*}{BT} & Clean & 88.97 & 89.52 & 81.02 & 91.77 & 91.97 & 85.14 & 89.10 & 89.54 & 81.07 & 89.46 & 89.90 & 81.70 \\
 & Perturbed & 64.40 & 69.80 & 53.60 & 72.40 & 69.50 & 53.30 & 71.10 & 56.30 & 39.20 & 66.60 & 37.70 & 23.20 \\
\midrule
\multirow{2}{*}{NI} & Clean & 89.37 & 89.91 & 81.67 & 92.41 & 92.16 & 85.46 & 88.38 & 89.70 & 81.32 & 88.41 & 87.64 & 78.00 \\
 & Perturbed & 74.80 & 75.20 & 60.20 & 72.60 & 69.70 & 53.50 & 77.90 & 69.30 & 53.00 & 73.50 & 65.20 & 48.40 \\
\midrule
\multirow{2}{*}{SSN} & Clean & 89.03 & 89.55 & 81.08 & 91.89 & 92.44 & 85.94 & 88.25 & 89.77 & 81.44 & 84.77 & 85.93 & 75.34 \\
 & Perturbed & 73.90 & 74.30 & 59.10 & 70.80 & 69.30 & 53.00 & 76.10 & 67.20 & 50.60 & 72.40 & 64.70 & 47.80 \\
\midrule
\multirow{2}{*}{GM} & Clean & 88.30 & 89.38 & 80.79 & 91.84 & 92.41 & 85.89 & 88.60 & 89.35 & 80.75 & 90.82 & 89.15 & 80.42 \\
 & Perturbed & 72.10 & 73.50 & 58.20 & 76.50 & 73.80 & 58.60 & 75.30 & 66.20 & 49.50 & 72.60 & 65.10 & 48.30 \\
\midrule
\multirow{2}{*}{Mixup} & Clean & 90.56 & 90.06 & 81.92 & 91.29 & 92.22 & 85.56 & 91.36 & 90.16 & 82.08 & 90.35 & 88.50 & 79.37 \\
 & Perturbed & 76.80 & 75.20 & 60.20 & 77.30 & 74.20 & 59.00 & 77.60 & 70.60 & 54.60 & 76.40 & 70.30 & 54.20 \\
\midrule
\multirow{2}{*}{ChatIE} & Clean & 62.92 & 50.20 & 33.50 & 67.75 & 47.30 & 31.00 & 62.40 & 62.60 & 45.60 & 66.70 & 34.60 & 20.90 \\
 & Perturbed & 59.30 & 42.30 & 26.80 & 63.40 & 32.10 & 19.10 & 64.40 & 69.80 & 53.60 & 69.20 & 55.70 & 38.60 \\
 \midrule
\multirow{2}{*}{SSDAU} & Clean & 92.00 & 92.05 & 85.27 & 92.80 & 92.95 & 86.83 & 91.74 & 92.90 & 86.74 & 91.58 & 89.94 & 81.77 \\
 & Perturbed & 83.20 & 80.70 & 67.70 & 88.10 & 89.80 & 81.40 & 80.10 & 77.90 & 63.90 & 84.50 & 86.40 & 76.10 \\
\bottomrule
\end{tabular}
\end{table*}

% \vspace{-0.2cm}
\paragraph{Protocol}
% hjw We evaluate没有说完
We select five models for three JERE task types: Multi-module Multi-Step (PRGC \citep{zheng2021prgc}, CasRel \citep{wei2020novel}), Multi-module One-Step (TPLinker \citep{wang2020tplinker}, SPN4RE \citep{sui2020joint}), and One-module One-Step (OneRel) \citep{OneRel}. 

We use the following metrics to evaluate {\toolname}: precision (Prec), F1-score (F1), and Intersection over Union (IoU). We also conduct significance tests to verify the significance of our improvements over other methods. All thresholds are selected on the validation set and fixed for test evaluation. Unless otherwise stated, all reported numbers are averaged over five random seeds.

\paragraph{Implementation} 
We conducted all experiments on a single server equipped with an Intel Xeon Gold 6248 2.50GHz CPU, two Tesla V100 SXM2 32GB GPUs, and Ubuntu 18.04.6 operating system. We reused the pre-trained BERT model (base-cased English) from Huggingface. 
For the consistency scoring module, we initialize $W$ and $b$ with pretrained contextual representations.

%Table 1

% \vspace{-0.2cm}

\paragraph{Dataset}
We conduct experiments on two representative English datasets, NYT~\citep{sandhaus2008new} and WebNLG \citep{gardent2017creating}. Both have two variations: fully annotated (NYT, WebNLG) and partially annotated (NYT$^{*}$, WebNLG$^{*}$). To evaluate robustness under semantic perturbation, we construct controlled evaluation datasets by injecting out-of-domain triples from SciER \citep{zhang2024scier} into all datasets at a 10\% rate. These triples are added as perturbed training instances rather than replacing original samples, creating a controlled setting with semantically related but out-of-domain structures.

\paragraph{Evaluation and Selection of Thresholds}
Table \ref{Epsilon} reports the number of augmented samples under different similarity thresholds. Across the four datasets, the number generally decreases as the threshold increases, and we select $\varepsilon=0.7$ on the validation set for all main experiments.

% \vspace{-0.3cm}
\subsection{Results}

% \vspace{-0.1cm}
\paragraph{Comparison with Baselines}
Table \ref{combined_results} presents the Prec., F1, and IoU results of {\toolname} and seven baselines on four dataset settings. Overall, {\toolname} achieves the best average performance among the compared methods in our evaluation and remains comparatively stable under semantically perturbed data. For example, on NYT, SSDAU shows a smaller F1 degradation than several conventional augmentation baselines under perturbation. These results suggest that explicitly constrained replacement with consistency filtering is better suited to JERE than unconstrained perturbation, especially when semantic ambiguity is present.

% Table \ref{compare_baseline} presents the results of SSADU with six baselines under the CasRel model.
Compared with Back Translation and Generative Models, preserving semantic structure is more effective for JERE than preserving semantic continuity. Compared with Noise Introduction and Same-tag Semantic Noise, tag-based discretization performs better than directly injecting noise. Likewise, compared with Word Substitution and Mixup, labeled discrete texts are more effective for JERE augmentation than unlabeled samples. {\toolname} also achieves higher precision and F1 than ChatIE across datasets. Overall, these results suggest that structure-preserving augmentation is better suited to precise JERE extraction than unconstrained perturbation, especially in ambiguous settings.

%the method of data augmentation by preserving the structured semantics of the text is superior to existing data augmentation strategies.

% Figure \ref{data_information2} displays the training results of {\toolname} with six baselines under the CasRel model at various iterations throughout the training process.
% The results reveal that {\toolname} consistently achieves optimal performance across different iteration numbers.
% These findings suggest that the augmented data produced by {\toolname} is beneficial for enhancing JERE models.
% Additionally, {\toolname} consistently delivers promising enhancements on four datasets in contrast to traditional methods.

% Figure 3
% \begin{figure*}[t]
% \subfigure[NYT$^{*}$] {
%  \label{nyt_star}
% \includegraphics[width=0.23\linewidth]{figs/draw_picture_NYT_star.pdf}
% }
% \subfigure[WebNLG$^{*}$] {
%  \label{webnlg}
% \includegraphics[width=0.23\linewidth]{figs/draw_picture_WebNLG_star.pdf}
% }
% \subfigure[NYT] {
%  \label{nyt}
% \includegraphics[width=0.23\linewidth]{figs/draw_picture_NYT.pdf}
% }
% \subfigure[WebNLG] {
%  \label{webnlg}
% \includegraphics[width=0.23\linewidth]{figs/draw_picture_WebNLG.pdf}
% }
%   \vspace{-3mm}
%   \caption{The effects of various baselines on different datasets are examined. Specifically, (a), (b), (c), and (d) illustrate the precision of the data augmentation baseline at different iterations, with the CasRel model serving as the prediction model.}
%   \label{data_information2}
% \end{figure*}

%Table 3
\begin{table*}[t]
\caption{Performance of different JERE models on four datasets. Each cell (A/B) denotes training on the original data (A) or the data augmented by {\toolname} (B). Bold values indicate improvements over the original data.} 
\label{result_five_model}
\footnotesize
\centering
\setlength{\tabcolsep}{5pt}
% \tabcolsep=0.15cm
% \vspace{0.4em} \centering%  xiaowu

\begin{tabular}{lcccc}%{\textwidth}
%\toprule[0.001pt]
\toprule
% \rule{0pt}{10pt} %此命令用于更改行高，
% \vspace{0.05cm}
    \multirow{1}{*}{Model}&\multicolumn{1}{c}{NYT$^{\ast}$}&\multicolumn{1}{c}{WebNLG$^{\ast}$}&\multicolumn{1}{c}{NYT}&\multicolumn{1}{c}{WebNLG}\\
   % &Pre.&Pre.&Pre.&Pre.\\
%\midrule[0.001pt]
\midrule
 SPN \citep{sui2020joint} & 91.44/\textbf{91.95} & 93.81/\textbf{96.84} & 92.67/92.64 & 90.21/\textbf{90.88} \\
 PRGC \citep{zheng2021prgc} & 93.33/\textbf{93.36} & 94.00/\textbf{94.46} & 93.54/\textbf{94.40} & 89.92/\textbf{91.32}\\
 CasRel \citep{wei2020novel} & 88.97/\textbf{91.47} & 91.77/\textbf{92.13} & 89.10/\textbf{91.74} & 89.46/\textbf{91.58} \\
 OneRel \citep{OneRel} & 90.17/\textbf{92.00} & 90.62/\textbf{92.80} & 92.83/\textbf{92.90} & 90.66/\textbf{91.60} \\
 TPLinker \citep{wang2020tplinker} & 90.23/{92.21} & 90.89/\textbf{91.34} & 91.33/\textbf{92.27} & 89.12/\textbf{89.93} \\
%\bottomrule[0.001pt]
\bottomrule
\end{tabular}
% \vspace{-10pt}
\end{table*}

\begin{table}[t]
\caption{Additional evaluation results in F1 (\%). ``Main'' reports the average over five random seeds, ``Budget'' uses equal augmentation size, and ``Low'' uses 10\% training data. The reported $p$-value corresponds to the average comparison in the Main setting.}
\label{tab:add_eval}
\centering
\footnotesize
\setlength{\tabcolsep}{4pt}
\begin{tabular}{lcccc}
\toprule
Method & Main & Budget & Low & $p$ \\
\midrule
Original & 90.67 & -- & 75.83 & -- \\
Mixup    & 90.21 & 90.04 & 77.18 & -- \\
GM       & 90.07 & 89.95 & 77.17 & -- \\
LLM Aug. & 87.89 & 87.42 & 74.63 & -- \\
SSDAU    & \textbf{91.95} & \textbf{91.55} & \textbf{79.75} & \textbf{0.019} \\
\bottomrule
\end{tabular}
\end{table}

We further evaluate robustness under semantic perturbation using data constructed with SciER \cite{zhang2024scier}. Table \ref{combined_results} compares SSDAU with seven baselines under clean and perturbed settings. Conventional augmentation baselines generally show larger performance degradation under perturbation, whereas SSDAU remains comparatively more stable across datasets under our evaluation protocol. This trend is consistent with the design of the consistency filtering module, which removes candidate replacements that are incompatible with the source semantics.

\vspace{-0.15cm}
\paragraph{Performance on Different JERE Tasks} 
Table \ref{result_five_model} shows the performance of different JERE models trained on the original data and data augmented by {\toolname}. The augmented data improve performance for most model--dataset combinations, although gains are not uniform across architectures. For example, the largest gain is observed on WebNLG$^{\ast}$ for SPN, while the change on NYT for SPN is negligible. These results indicate that the usefulness of augmentation depends on the underlying JERE architecture and dataset annotation setting. We also observe larger gains on partially annotated datasets than on fully annotated ones, suggesting that structured augmentation may be especially helpful when supervision is less complete.

\paragraph{Additional evaluation.}
Table~\ref{tab:add_eval} summarizes results on stability, fairness, and reduced-data settings. {\toolname} remains the best method on average under five random seeds, equal augmentation budgets, and 10\% training data. The corresponding $p$-value in the table is computed for the average comparison under this setting.

% \vspace{-0.35cm}
\subsection{Ablation Study}
% \vspace{-0.1cm}

%Table 4
\begin{table}[t]
\caption{Ablation study for {\toolname}. ``No Split'' denotes not splitting the text. ``No Label Split'' denotes splitting by semantics without semantic tag. ``Full Split'' denotes complete splitting of the words in the text.} 
\label{DDR}
\centering
\setlength{\tabcolsep}{3pt}
\footnotesize
% \resizebox{70mm}{28mm}{
\begin{tabular}{lccc}
\toprule
% \rule{0pt}{10pt} 
% \vspace{0.05cm}
    Dataset&NYT$^{\ast}$&WebNLG$^{\ast}$&Avg.\\
\midrule
   CasRel Baseline&90.17&90.62&90.39\\
   {\toolname}&92.00&92.80&92.40\\
\midrule
 \textit{Ablation for Pre-processing}\\
 No Split&89.32&90.17&89.75\\
 No Label Split&90.33&90.42&90.38\\
 Full Split&88.64&89.76&89.20\\
\midrule
 \textit{Ablation for Augmentation}\\
 (h,t)&64.21&73.83&69.02\\
 (r)&77.42&84.31&80.87\\
 (h,r,t)&90.41&91.13&90.77\\
 (h,r,h)&85.66&88.53&87.10\\
 (t,r,t)&82.12&84.44&83.28\\
\midrule
 \textit{Ablation for Filtering}\\
 No Filtering&89.92&90.84&90.38\\
%\bottomrule[0.001pt]
\bottomrule
\end{tabular}
% }

\end{table}

\begin{table*}[t]
\caption{Semantic consistency verification of augmented text. $\nu \in [0,1]$ denotes normalized syntactic coherence, where larger values indicate better agreement with the source syntactic pattern.} 
%\vspace{\baselineskip}
\label{verfication_syntactic}
\footnotesize
\centering
% \vspace{0.4em} \centering%  xiaowu
\begin{tabular}{lc}%{\textwidth}
%\toprule[0.001pt]
\toprule
    \multirow{3}{*}{Source}&\multicolumn{1}{l}{\begin{tabular}[c]{@{}l@{}}Text = South Africa, and the rest of Africa.
    \end{tabular}}\\
   &\multicolumn{1}{l}{\begin{tabular}[c]{@{}l@{}}Triple = [[Africa, /location/location/contains, South Africa]] \\ 
    \end{tabular}}\\
    &\multicolumn{1}{l}{\begin{tabular}[c]{@{}l@{}}Structured Semantic = location contain location \\ 
    \end{tabular}}\\
%\midrule[0.001pt]
\midrule
\multirow{4}{*}{Syntax Matching}&\multicolumn{1}{l}{\begin{tabular}[c]{@{}l@{}}Text1 = South Africa is a part of Africa. \textcolor{red}{$\nu$ = 0.516}
    \end{tabular}}\\
    &\multicolumn{1}{l}{\begin{tabular}[c]{@{}l@{}}Text2 = North Africa, and the rest of Africa. \textcolor{blue}{$\nu$ = 0.923}
    \end{tabular}}\\
    
   &\multicolumn{1}{l}{\begin{tabular}[c]{@{}l@{}}Triple = [[Africa, /location/location/contains, North Africa]] \\ 
    \end{tabular}}\\
    &\multicolumn{1}{l}{\begin{tabular}[c]{@{}l@{}} Structured Semantic = location contain location \\ 
    \end{tabular}}\\
 \bottomrule
\end{tabular}
% \vspace{-10pt}
\end{table*}

We conduct ablation studies on the main components of {\toolname} under consistent settings.

% \vspace{-0.2cm}
\paragraph{Data Discretization and Reconstruction}
We evaluate performance after removing the pre-processing component by directly splitting data based on triad messages without semantic tags, and by applying conventional \textit{no-split} and \textit{full-split} schemes \citep{gao2020split}.

As shown in Table \ref{DDR}, we evaluate the effectiveness of the pre-processing component using precision as the metric. {\toolname} outperforms \textit{No Split}, \textit{No Label Split}, and \textit{Full Split}, showing that preserving semantic labels and span boundaries is important for structured augmentation in JERE.

% First, we remove the pre-processing component: Data Discretization and Reconstruction.  
% We directly split the data based on the triad message without semantic tags (No Label Split).
% Besides, we employ conventional text split methods: the no split and complete full split schemes \citep{gao2020split}. 
% As shown in Table \ref{DDR}, we evaluate the effectiveness of pre-processing components before and after removal by precision. 
% Our Data Discretization and Reconstruction is shown to perform better than without pre-processing component (about 2.02\%-3.20\% improvement).
% Furthermore, the semantic tagging prompts will positively impact discrete text data augmentation under the low-resource JERE tasks.

% \vspace{-0.2cm}
\paragraph{Structured Semantic Data Augmentation} 
We evaluate the augmentation component by varying which parts of the triple are replaced after preprocessing. As shown in Table \ref{DDR}, only coordinated replacement of $(h, r, t)$ yields a positive effect, while partial replacement strategies such as $(h,t)$ or $(r)$ substantially degrade performance. This indicates that structured replacement is not sufficient: when semantic constraints are only partially enforced, the generated data can become harmful. The result also explains why SSDAU is designed as a pipeline rather than a single replacement step.

\paragraph{Scoring-based Consistency Filtering}
We assess the impact of the consistency filtering component in {\toolname}. Table \ref{DDR} shows that removing filtering reduces performance to below the baseline trained on the original data only. This result indicates that structured replacement can introduce noisy candidates if applied without quality control, and that consistency filtering is therefore necessary for obtaining reliable gains from augmentation.

\begin{table*}[t]
\caption{Ablation study on parameter initialization across four datasets.}
\small
\label{ablation_init}
\centering
\setlength{\tabcolsep}{4pt}
\begin{tabular}{lcccccccccccc}
\toprule
    \multirow{3}{*}{Init Type}&\multicolumn{6}{c}{Partial Match}&\multicolumn{6}{c}{Exact Match}\\
    \cmidrule(r){2-7}
    \cmidrule(r){8-13}
    \ &\multicolumn{3}{c}{NYT$^{\ast}$}&\multicolumn{3}{c}{WebNLG$^{\ast}$}&\multicolumn{3}{c}{NYT}&\multicolumn{3}{c}{WebNLG}\\
    \cmidrule(r){2-4}
    \cmidrule(r){5-7}
    \cmidrule(r){8-10}
    \cmidrule(r){11-13}
   &Prec.&F1&IoU&Prec.&F1&IoU&Prec.&F1&IoU&Prec.&F1&IoU\\
\midrule
 Pretrained & 92.00 & 92.05 & 85.27 & 92.80 & 92.95 & 86.83 & 91.74 & 92.90 & 86.74 & 91.58 & 89.94 & 81.77 \\
 Random & 73.92 & 52.30 & 35.41 & 76.54 & 65.72 & 48.97 & 67.35 & 48.12 & 31.69 & 90.26 & 62.45 & 45.36 \\
 Zero & 78.42 & 70.83 & 54.85 & 83.78 & 54.97 & 37.90 & 78.22 & 70.43 & 54.42 & 87.65 & 51.78 & 34.94 \\
\bottomrule
\end{tabular}
\end{table*}

\begin{table*}[htb]
\caption{Augmented data generated by {\toolname}. Black texts are the original examples. \textcolor{red}{Red texts} are the discrete text. \textcolor{blue}{Blue texts} are the precondition for text segmentation and augmentation. $\varepsilon_{1}$ is the entity similarity threshold and $\varepsilon_{2}$ is the relation similarity threshold.} 
%\vspace{\baselineskip}
\footnotesize
\label{case study}
\centering
% \vspace{0.4em} \centering%  xiaowu
\begin{tabular}{cc}%{\textwidth}
%\toprule[0.001pt]
\toprule
    \multirow{3}{*}{Source}&\multicolumn{1}{l}{\begin{tabular}[c]{@{}l@{}}Text: At Arkansas , the freshman Mitch Mustain led the Razorbacks in a 24-23 double- \\ overtime upset of Alabama.
    \end{tabular}}\\
   &\multicolumn{1}{l}{\begin{tabular}[c]{@{}l@{}}Triples: Mitch Mustain(people)$|$Arkansas(place)$|$place\_lived \\ 
\quad \quad \quad \ \ Razorbacks(group)$|$Mitch Mustain(people)$|$contain
    \end{tabular}}\\
%\midrule[0.001pt]
\midrule
 \multirow{5}{*}{\begin{tabular}[c]{@{}l@{}}Head \\ \ \ $\rightarrow$ \\ Head \end{tabular}}&\multicolumn{1}{l}{\begin{tabular}[c]{@{}l@{}}Condition: \textcolor{blue}{$Tag_{h} = people, Tag_{t} = place, Tag_{r} = place\_lived, \Theta_{h} \geq \varepsilon_{1}$}.
    \end{tabular}}\\
    &\multicolumn{1}{l}{\begin{tabular}[c]{@{}l@{}}Text: At Arkansas, the freshman \textcolor{red}{Amy Grant} led the Razorbacks in a 24-23 double-ov- \\ ertime upset of Alabama.
    \end{tabular}}\\
   &\multicolumn{1}{l}{\begin{tabular}[c]{@{}l@{}}Triples: \textcolor{red}{Amy Grant}(people)$|$Arkansas(place)$|$place\_lived \\ 
\quad \quad \quad \ \ Razorbacks(group)$|$\textcolor{red}{Amy Grant}(people)$|$contain
    \end{tabular}}\\
 \midrule
 \multirow{5}{*}{\begin{tabular}[c]{@{}l@{}}Tail \\ \ $\rightarrow$ \\ Tail \end{tabular}}&\multicolumn{1}{l}{\begin{tabular}[c]{@{}l@{}}Condition: \textcolor{blue}{$Tag_{h} = people, Tag_{t} = place, Tag_{r} = place\_lived, \Theta_{t} \geq \varepsilon_{1}$}.
    \end{tabular}}\\
    &\multicolumn{1}{l}{\begin{tabular}[c]{@{}l@{}}Text: At \textcolor{red}{Nashville}, the freshman Mitch Mustain led the Razorbacks in a 24-23 double- \\ overtime upset of Alabama.
    \end{tabular}}\\
   &\multicolumn{1}{l}{\begin{tabular}[c]{@{}l@{}}Triples: Mitch Mustain(people)$|$\textcolor{red}{Nashville}(place)$|$place\_lived \\ 
\quad \quad \quad \ \ Razorbacks(group)$|$Mitch Mustain(people)$|$contain
    \end{tabular}}\\
 \midrule
 \multirow{5}{*}{\begin{tabular}[c]{@{}l@{}}Relation \\ \ \ \ \ \ $\rightarrow$ \\ Relation \end{tabular}}&\multicolumn{1}{l}{\begin{tabular}[c]{@{}l@{}}Condition: \textcolor{blue}{$Tag_{h} = people, Tag_{t} = place, \Theta_{r} \geq \varepsilon_{2}$}.
    \end{tabular}}\\
    &\multicolumn{1}{l}{\begin{tabular}[c]{@{}l@{}}Text: At Arkansas, \textcolor{red}{the freshman} Mitch Mustain \textcolor{red}{led the} Razorbacks \textcolor{red}{in a 24-23 double-} \\ \textcolor{red}{overtime upset of Alabama}.
    \end{tabular}}\\ 
   &\multicolumn{1}{l}{\begin{tabular}[c]{@{}l@{}}Triples: Mitch Mustain(people)$|$Arkansas(place)$|$\textcolor{red}{location} \\ 
\quad \quad \quad \ \ Razorbacks(group)$|$Mitch Mustain(people)$|$contain
    \end{tabular}}\\
 \bottomrule
\end{tabular}
% \vspace{-10pt}
\end{table*}

% Finally, we verify the impact of the consistency filtering component in SSDAU. Table \ref{DDR} shows the precision of the JERE models without any filtered data compared to the models with filtered data. 
% The result shows that filter data positively impact the model's precision, which decreases when low-quality augmented data are not removed, implying that consistent filtering is crucial to the precision of the model.

\paragraph{Parameter Initialization}
We investigate the impact of parameter initialization by comparing three methods: random initialization, zero initialization, and pretrained initialization (used in SSDAU). As shown in Table \ref{ablation_init}, pretrained initialization consistently outperforms others across all four datasets. Significance tests confirm these improvements are statistically significant (p=0.012 on NYT, p=0.009 on WebNLG, p=0.016 on NYT$^{\ast}$, p=0.008 on WebNLG$^{\ast}$). These results validate our choice of pretrained initialization for consistency scoring.

\subsection{Analysis}

\paragraph{Semantic coherence analysis.}
During the semantic coherence analysis of {\toolname}, we follow a two-step process to assess whether the augmented text preserves structural compatibility with the source triples. First, we augment texts by considering similarities between annotations of the same type while preserving the semantic tag pattern (e.g., ``location contains location''). Next, we use Biber Tagger \cite{sharoff21rs} to compare syntactic patterns between source and augmented texts. Table \ref{verfication_syntactic} reports two examples with different syntactic coherence scores $\nu$, where larger $\nu$ indicates better agreement with the source syntactic pattern. We discard low-coherence candidates during filtering and keep only those that satisfy the relevance threshold.

\paragraph{Training Cost and Convergence.}

\begin{figure}[t]
\subfigure[The Partially Data] {
 \label{partial_data}
\includegraphics[width=\linewidth, height= 0.44\linewidth]{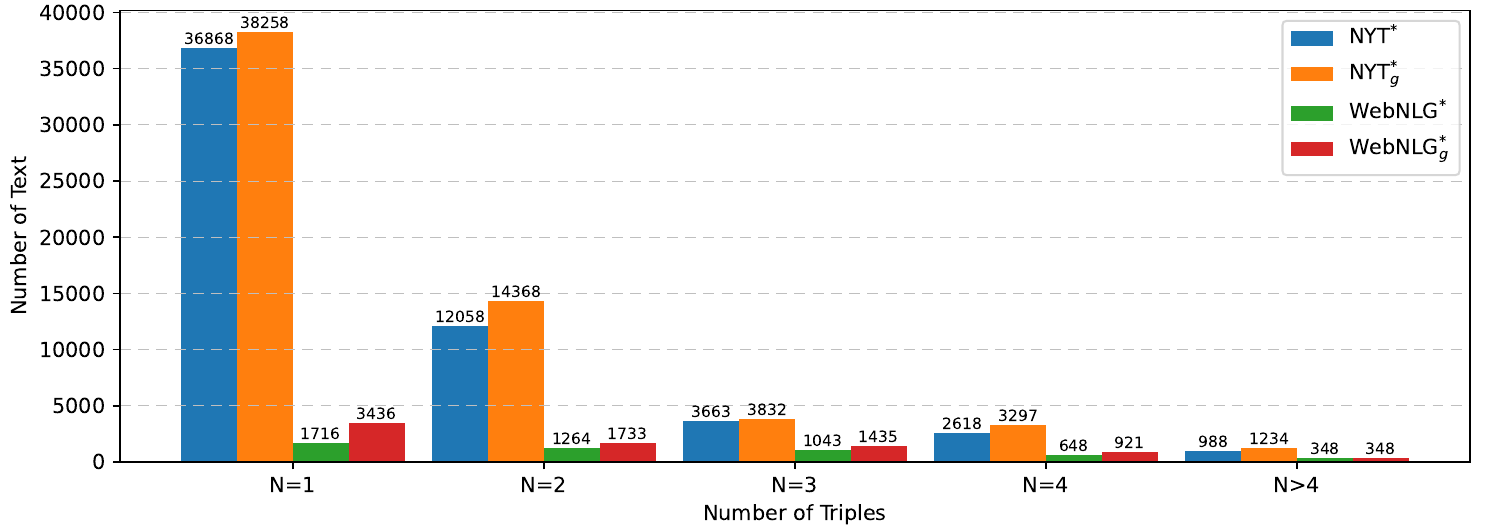}
}
\vfill
\subfigure[The Exactly Data] {
 \label{exact_data}
\includegraphics[width=\linewidth, height= 0.44\linewidth]{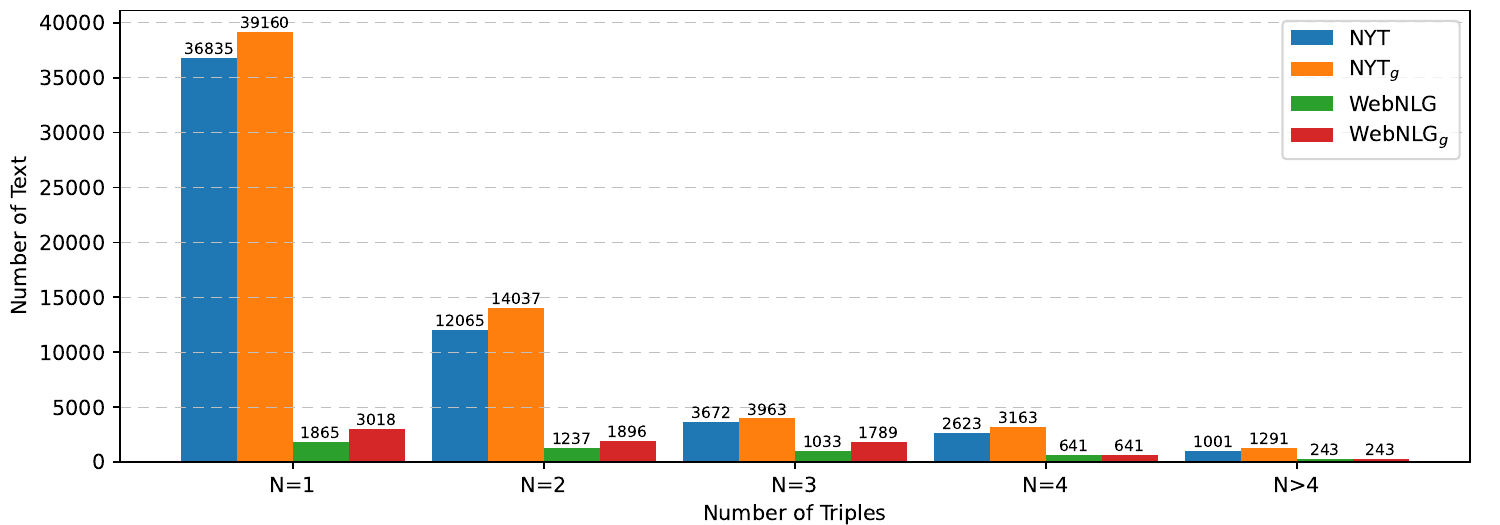}
}
  
  \caption{The comparison between the number of text after augmentation with {\toolname} and the initial one for different types of datasets.}
  \label{fig:comparison_partial}
  
\end{figure}
% \vspace{-0.5cm}

Figure \ref{fig:comparison_partial} provides details about the original and augmented texts containing varying numbers of triplets. We focus on scenarios where an entity appears in multiple triplet relations and categorize texts by triplet number to evaluate the effectiveness of {\toolname} for such texts. By classifying the augmented data by triplet count and incorporating it into the training set, we assess the performance of different JERE models using the same test set. The results demonstrate the effectiveness of {\toolname} for texts with different triplet counts. Our method proves valuable across texts with varying numbers of triplets, showing that as the number of triplets in the training set decreases, the availability of augmented data increases, leading to improved model precision.

\subsection{Case Study}
Table \ref{case study} presents three cases of {\toolname} applied to JERE tasks.
Our augmentation approach increases structural diversity while preserving the target entity and relation configuration. At the same time, these examples also illustrate an important limitation of replacement-based augmentation: an augmented sentence can remain structurally valid for JERE while being factually implausible in the real world. In SSDAU, this trade-off is partially controlled by semantic matching and consistency filtering, but not completely eliminated.

% \vspace{-0.1cm}
\section{Conclusion}
We propose {\toolname}, a structured data augmentation pipeline for JERE that combines semantic segmentation, constrained replacement, and consistency filtering. Our approach integrates contextualized embeddings with traditional similarity scores to better distinguish semantically similar but non-equivalent entities, and uses topic-aware filtering to control noisy augmentations. Experiments on NYT and WebNLG show that {\toolname} improves average extraction performance in most settings and yields better robustness under semantic perturbation. The ablation results further indicate that the benefits come from the interaction between structure-aware replacement and consistency filtering, with filtering playing an important role in quality control. These findings suggest that explicitly constrained augmentation remains a useful direction for JERE.

% \vspace{-0.1cm}
\section*{Limitations}
Although {\toolname} improves average performance over the compared baselines in our experiments, it has several limitations. 
First, as a replacement-based augmentation method, {\toolname} cannot introduce entirely new entity types or unseen relation schemas. 
Second, the method assumes access to entity annotations or type information for segmentation and constrained replacement, which may limit its direct applicability to settings without such supervision.
Third, some augmented sentences may remain structurally valid for JERE while being factually implausible, which may be undesirable for downstream applications such as knowledge graph construction.
Fourth, the filtering component becomes particularly important when the candidate augmentation pool is noisy; as shown in our ablation, removing filtering can eliminate the gains from augmentation.
Finally, our experiments focus on English NYT/WebNLG benchmarks, and we do not evaluate transfer to multilingual, nested-entity, or domain-specific IE settings.

\bibliography{reference}

\end{document}